\documentclass{article}

\usepackage[preprint]{neurips_2024}

\usepackage[utf8]{inputenc} 
\usepackage[T1]{fontenc}    
\usepackage{hyperref}       
\usepackage{url}            
\usepackage{booktabs}       
\usepackage{amsfonts}       
\usepackage{nicefrac}       
\usepackage{microtype}      
\usepackage{xspace}
\usepackage{graphicx}
\usepackage{caption}
\usepackage{amsmath}
\usepackage[table]{xcolor}

\usepackage{multirow}
\usepackage{multicol}
\usepackage{wrapfig}
\usepackage{adjustbox}

\def\Ours{X-Omni\xspace}

\title{\Ours: Reinforcement Learning Makes Discrete Autoregressive Image Generative Models Great Again}

\author{%
  \textbf{Zigang Geng$^\ast$} \quad
  \textbf{Yibing Wang$^\ast$} \quad
  \textbf{Yeyao Ma} \quad
  \textbf{Chen Li} \quad
  \textbf{Yongming Rao} \hspace{1.0em}\\[2.2mm]
  \hspace{0.1em} \textbf{Shuyang Gu} \quad
  \textbf{Zhao Zhong} \quad
  \textbf{Qinglin Lu} \quad
  \textbf{Han Hu$^\ddag$} \quad
  \textbf{Xiaosong Zhang$^\dag$}\\[2.2mm]
  \textbf{Linus} \quad
  \textbf{Di Wang} \quad
  \textbf{Jie Jiang} \\[3mm]
  Tencent Hunyuan X \\[2mm]
  {\fontsize{9.4pt}{9.84pt}\selectfont  \textsuperscript{$\ast$}Equal contribution ~~~\textsuperscript{$\dag$}Project Lead ~~~\textsuperscript{$\ddag$}Corresponding Author} \\[2mm]
  \url{https://x-omni-team.github.io}
}

\begin{document}

\makeatletter
\let\@oldmaketitle\@maketitle
\renewcommand{\@maketitle}{\@oldmaketitle
    \centering
    \vspace{-15pt}
    \adjustbox{width=0.95\linewidth,center}{
        \includegraphics{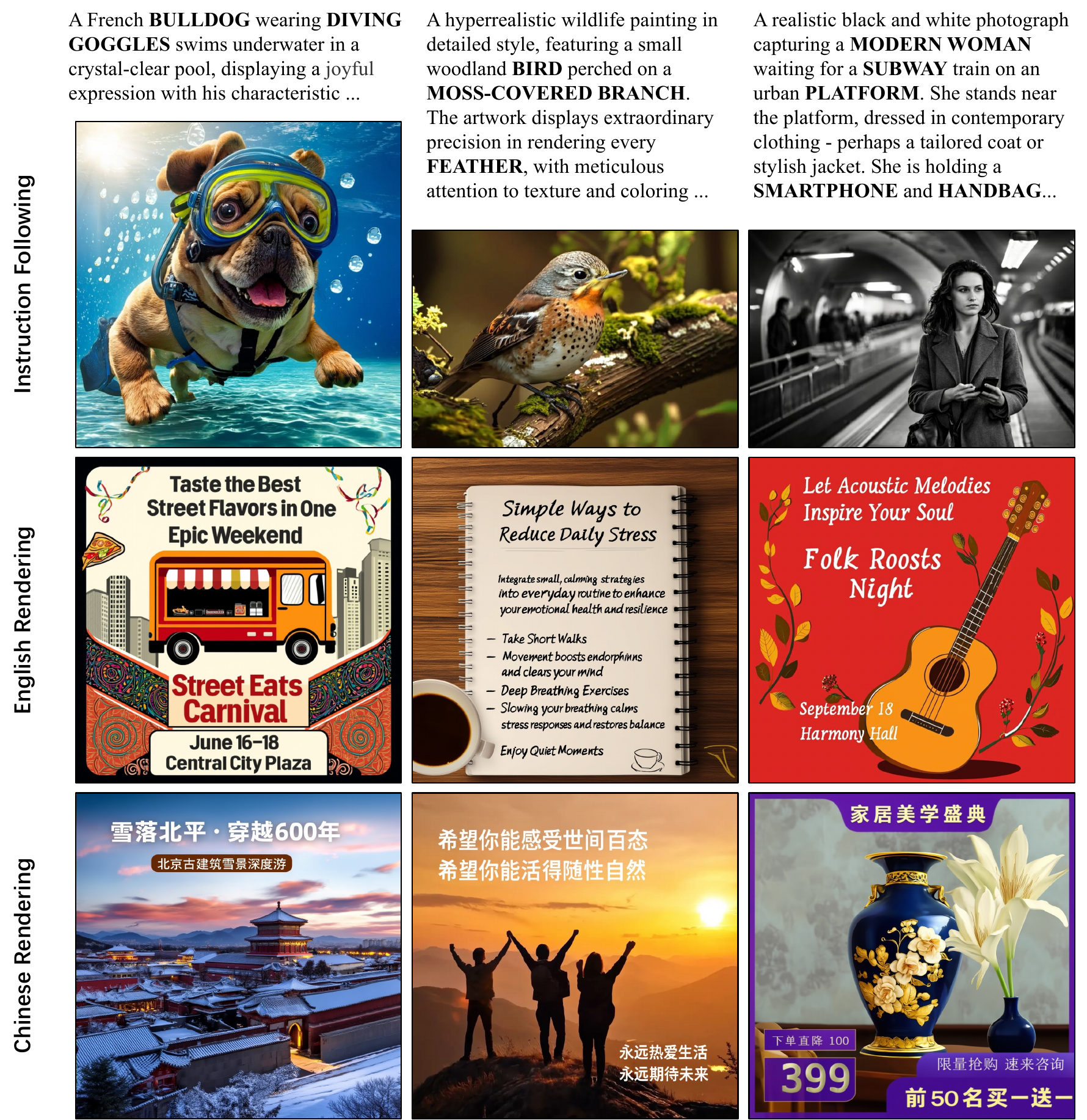}
    }
    \captionof{figure}{Equipped with reinforcement learning, \Ours integrates image and language modeling in a unified autoregressive framework, enabling it to generate high-quality images following complex instructions and render long text in both English and Chinese.}
    \label{fig:teaser}
    \bigskip
}   
\makeatother

\maketitle

\begin{abstract}

Numerous efforts have been made to extend the ``next token prediction'' paradigm to visual contents, aiming to create a unified approach for both image generation and understanding. Nevertheless, attempts to generate images through autoregressive modeling with discrete tokens have been plagued by issues such as low visual fidelity, distorted outputs, and failure to adhere to complex instructions when rendering intricate details. These shortcomings are likely attributed to cumulative errors during autoregressive inference or information loss incurred during the discretization process.
Probably due to this challenge, recent research has increasingly shifted toward jointly training image generation with diffusion objectives and language generation with autoregressive objectives, moving away from unified modeling approaches. In this work, we demonstrate that reinforcement learning can effectively mitigate artifacts and largely enhance the generation quality of a discrete autoregressive modeling method, thereby enabling seamless integration of image and language generation. Our framework comprises a semantic image tokenizer, a unified autoregressive model for both language and images, and an offline diffusion decoder for image generation, termed \Ours. \Ours achieves state-of-the-art performance in image generation tasks using a 7B language model, producing images with high aesthetic quality while exhibiting strong capabilities in following instructions and rendering long texts.

\end{abstract}

\section{Introduction}

The ``next token prediction'' paradigm in language modeling has catalyzed the ongoing AI revolution \cite{gpt3, chagpt, achiam2023gpt, liu2024deepseek, touvron2023llama, touvron2023llama2, dubey2024llama, qwen, yang2024qwen2}. This intuitiveness stems from language's inherent properties: tokens are discrete and sequentially structured. In the image domain, pioneering works like iGPT and DALL-E attempted to replicate this paradigm for image generation by discretizing images into sequential tokens and generating content through per-step token prediction. Notably, DALL-E demonstrated the feasibility of generating impressive images from text inputs.

However, the image quality produced by early DALL-E remained limited. Despite subsequent efforts to enhance quality—such as those in~\cite{sun2023emu, emu2, wang2024emu3, dong2023dreamllm, wu2024vila, team2024chameleon}—generated images still exhibited relatively low fidelity. This limitation is likely attributable to cumulative errors arising from the sequential generation of image tokens. It is presumably due to this issue that the field has experienced a swift transition toward the adoption of diffusion models for this task~\cite{ramesh2022hierarchical, rombach2022high, gu2022vector, saharia2022photorealistic}. Unfortunately, the architectural and modeling heterogeneity of these approaches presents challenges for integrating robust semantic capabilities into image generation. While various hybrid designs~\cite{qu2025tokenflow, chen2025janus, chen2025semhitok, huang2025illume+} have been proposed to address this issue, the inherent cross-modal modeling mismatch remains a barrier, resulting in suboptimal solutions. Notably, recent research interest has shifted toward jointly training image generation with diffusion objectives and language generation with autoregressive objectives~\cite{xie2024show, zhou2024transfusion, kou2024orthus, chen2025blip3, pan2025transfer, lin2025uniworld, deng2025bagel, wu2025omnigen2, xie2025showo2}, which has further exacerbated this modeling mismatch.

In this work, we advocate for a discrete autoregressive framework for image generation that unifies the modeling of text and images, thereby facilitating better knowledge transfer and capability sharing across vision and language modalities. Our key insight is that reinforcement learning can significantly mitigate the limitations of autoregressive methods. When paired with carefully designed reward models, reinforcement learning enables automatic refinement of generative models — effectively reducing cumulative errors and producing tokens that align more closely with diffusion decoders.

\begin{figure}[t]
  \centering
  \includegraphics[width=\linewidth]{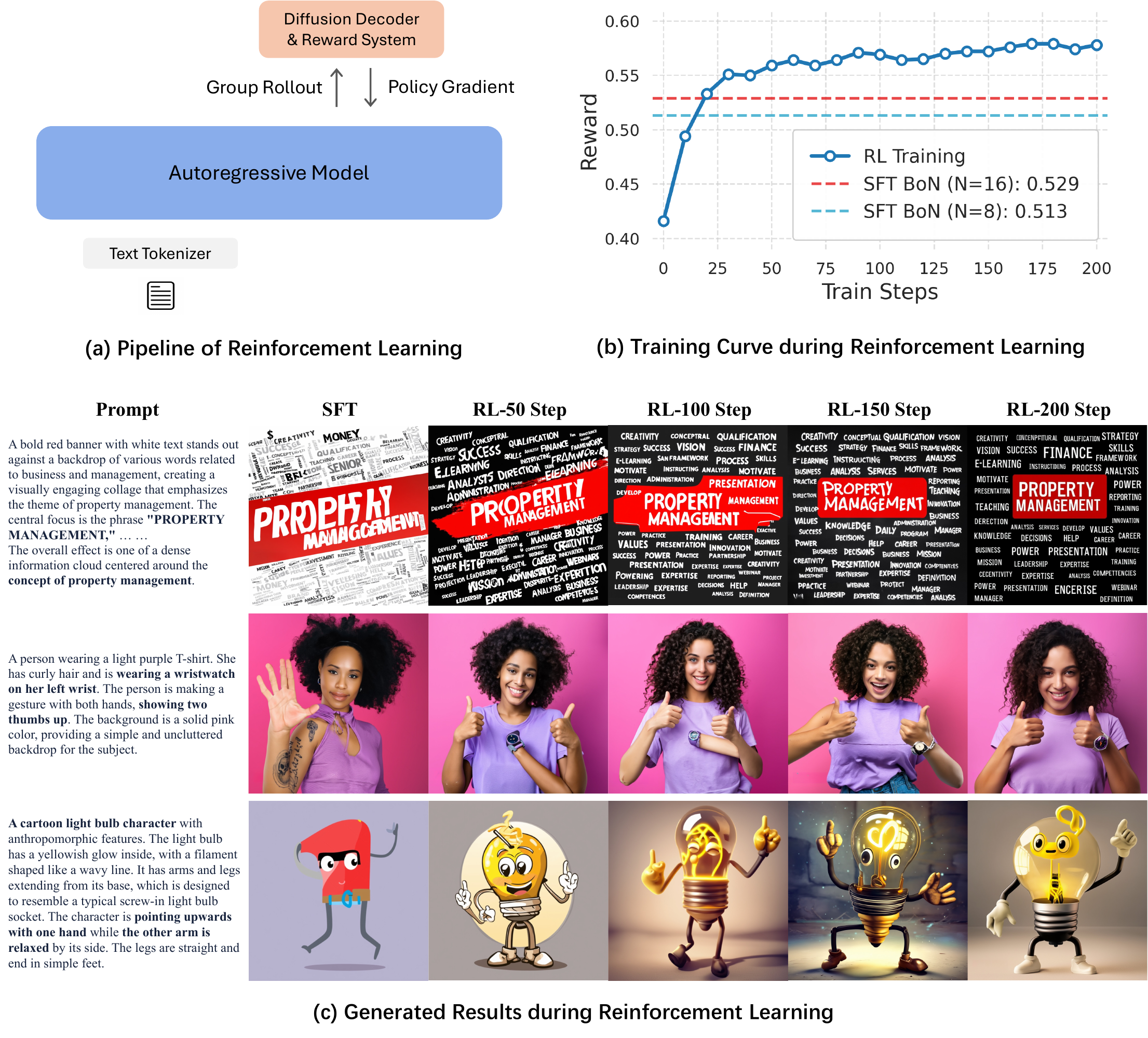}
  \vspace{-10pt}
  \caption{During the reinforcement learning process, \Ours's image generation reward quickly surpassed the best-of-N results achieved by the SFT model (SFT BoN) and demonstrated steady improvement. This progress was reflected in the model's text rendering capabilities, the aesthetic quality of the generated images, and its ability to follow instructions, all of which improved gradually.}
  \label{fig:2}
\end{figure}

As illustrated in Figure~\ref{fig:2}(c), the autoregressive generative models after supervised fine-tuning (SFT) produce relatively low-quality outputs, characterized by incorrect text generation, distorted body features, and failure to follow intricate instructions. During the reinforcement learning process, the aesthetic quality of generated images is gradually enhanced, and the ability to adhere to instructions and the capacity to render long texts improve steadily. Figure~\ref{fig:teaser} presents additional examples generated by our proposed method, which demonstrate high-quality visual appearances, robust capability in following complex instructions, and accurate rendering of long texts in both English and Chinese.

Another key feature of our design lies in the adoption of semantic encoders—such as SigLIP 2—to generate discrete tokens, a choice that also lends itself naturally to image understanding tasks. Combining with the isomorphic autoregressive modeling shared between image and text generation, we can seamlessly integrate the two core tasks of image generation and image understanding within a unified network. In contrast to recent works that rely on heterogeneous image encoders~\cite{chen2025janus,deng2025bagel,liao2025mogao}, our approach eliminates the need to re-extract semantic embeddings from generated images for understanding purposes during multi-turn joint image understanding and generation. This makes the overall architecture substantially more streamlined and efficient. We refer to this network as \Ours. Through this work, we aim to redirect research attention back to discrete autoregressive methods for image generation.

\section{Related Work}

\paragraph{Continuous tokens versus discrete tokens.}
Recent advancements in ``next-token prediction'' have showcased remarkable capabilities in both language and image understanding tasks, sparking significant interest in integrating image generation into this unified framework. Early approaches drew inspiration from the continuous visual representations employed in visual language models, either by autoregressively predicting continuous visual tokens~\cite{sun2023emu,emu2,tong2024metamorph} or by using query tokens to predict continuous visual tokens in parallel~\cite{wu2024next,ge2024seed}. These methods rely on MSE loss or cosine similarity to supervise the prediction of continuous visual tokens, relying on distributional assumptions like the Gaussian distribution. However, these assumptions limit the range of image distributions the methods can effectively represent. Furthermore, the means by which reinforcement learning can be integrated into continuous token methods remains unclear, potentially constraining their upper limits.

In contrast, autoregressive models utilizing discrete visual tokens can generate more complex distributions but grapple with challenges stemming from image discretization losses. For instance, Chameleon~\cite{team2024chameleon} and Emu3~\cite{wang2024emu3} rely on predicting discretized image tokens via lossy quantization, which constrains the level of detail in generated images. To mitigate this, some approaches~\cite{wu2024vila,qu2025tokenflow,wu2025janus,chen2025janus} attempt to introduce semantic supervision during the image pixel discretization process, aiming to enhance the model's capabilities in image understanding and generation. LaViT~\cite{jin2024unified} reconstructs the semantic features of images extracted by EVA-CLIP~\cite{sun2023eva} for discretization, followed by a diffusion decoder to incorporate finer details. While this method replaces image pixels with more refined semantic features for discretization, it encounters a distribution gap between the autoregressively generated image tokens and the ground truth tokens used in training the diffusion decoder. Our approach addresses these challenges by leveraging reinforcement learning to align the distribution of autoregressively generated image tokens with the expected distribution of the diffusion decoder, thereby enabling high-quality image generation through discrete autoregressive methods.

\paragraph{Different options to hybrid AR and diffusion models.}
In the literature, there are several approaches that combine autoregression with diffusion models.
One line of work~\cite{dong2023dreamllm,pan2025transfer,chen2025blip3} employs a serial architecture, where query tokens extract continuous conditions from the autoregressive model to guide the diffusion model’s generation process. Another more prevalent line~\cite{zhou2024transfusion,shi2024lmfusion,xie2024show,xie2025showo2,ma2025janusflow,xiao2024omnigen,wu2025omnigen2,liao2025mogao,deng2025bagel} adopts a parallel architecture: the autoregressive model (processing language tokens) and the diffusion model (processing visual latents) perform forward passes simultaneously, with information exchanged via self-attention layers. While these approaches can generate high-fidelity images due to their parallel and diffusion-based nature, they also exhibit a loose connection between the two components. Furthermore, reinforcement learning techniques for diffusion models are less mature compared to those for autoregressive models, limiting the application of reinforcement learning in optimizing these image generation models.

\paragraph{Reinforcement learning for generative models.}
Most reinforcement learning research for generative models has focused on diffusion models. For instance, \cite{clark24diff,xu23imagereward,mihir24videorg} fine-tune diffusion models using differentiable rewards via reward gradients. Other works~\cite{lee23align,fan23dpok} optimize through reward-weighted negative log-likelihood objectives. Additionally, some studies extend standard reinforcement learning approaches from language modeling to diffusion models~\cite{dong23raft,fan23dpok,geng24instructdiffusion,wallace24dpo,yang24dpo,liang24spo,black24dpo,miao24dpo,yuan24selfplay,hu25sparse,gupta25ppo,liu2025flowgrpotrainingflowmatching}, such as DPO~\cite{rafael23dpo}, PPO~\cite{schulman17ppo}, and GRPO~\cite{shao2024deepseekmath}. While these methods achieve certain effectiveness, their potential remains limited due to insufficient diversity in sampling. In contrast, our work aims to apply reinforcement learning algorithms to autoregressive generative models. Leveraging the maturity of reinforcement learning techniques for autoregressive methods, we demonstrate that such an approach can yield very strong results.

\section{Method}
\subsection{Overall Architecture}

\begin{wrapfigure}{r}{0.5\linewidth}
  \centering
  \vspace{-45pt}
  \includegraphics[width=0.9\linewidth]{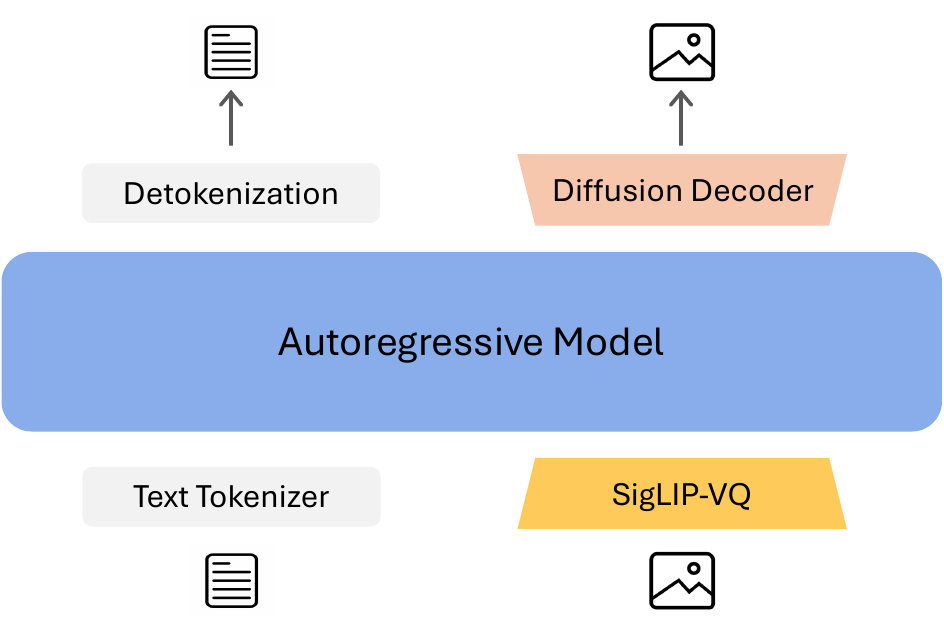}
  \caption{The architecture of \Ours.}
  \label{fig:3}
  \vspace{-10pt}
\end{wrapfigure}

We propose integrating image and text tokens within a unified autoregressive architecture. As illustrated in Figure~\ref{fig:3}, the framework is built with an autoregressive model, which incorporates a SigLIP-VQ tokenizer, and a diffusion decoder for the image generation. These components enable text-like tokenization and detokenization for both image and text inputs and outputs within a unified autoregressive framework. Detailed components are listed as follows.

\paragraph{Image Tokenization}
Instead of pixel-level reconstruction objectives, our image tokenizer is trained on visual understanding tasks. To convert continuous images into discrete tokens while retaining rich semantic information, we select a pre-trained SigLIP2-g~\cite{tschannen2025siglip} ViT as the visual semantic extractor. Building on the ViT encoder, we incorporate a vector quantizer as image tokenizer and align with a pre-trained large language model (LLM), Qwen2.5-1.5B~\cite{yang2024qwen2} on visual understanding tasks. The vector quantizer adopts a codebook of 16,384 vocabulary size and 2,048 dimension, with a residual block as an adapter between the visual tokenizer and the LLM. The visual encoder and vector quantizer collectively constitute the SigLIP-VQ image tokenizer. Both components are maintained frozen during the subsequent training phases, ensuring stability and consistency in the tokenization process. 

\paragraph{Autoregressive Modeling}
After tokenizing images into discrete tokens, both visual and language tokens can be naturally unified through multimodal modeling within an autoregressive architecture. \Ours adopts Qwen2.5-7B~\cite{yang2024qwen2} as the base pre-trained model. To integrate visual perception capabilities into the text-based language model, we insert four randomly initialized vision-specific blocks both before and after the original transformer layers. These vision-specific blocks are designed with the same structural configuration as standard transformer blocks, but operate exclusively on image tokens, leaving text tokens unaffected. Additionally, we introduce randomly initialized embedding layers and classification heads for image tokens. Compared to  the original language model, our architectural modifications introduce no extra infrastructure complexity while maintaining full compatibility with distributed training strategies, including tensor, pipeline and context parallelism.

For visual generation and understanding tasks, visual tokens and language tokens are concatenated into a unified multimodal sequence, which is then input to the autoregressive model for next-token prediction training. For understanding tasks, only the language tokens are supervised, whereas for generation tasks, only the visual tokens are supervised. Furthermore, to accommodate arbitrary image resolutions, we prepend a resolution information prefix to the visual tokens in the following format:
\[
\text{language tokens <SOM> height width <Image> visual tokens }\text{ <EOM> language tokens,}
\]
where the special tokens ``<SOM>'' and ``<EOM>'' denote the start and end markers within the multimodal sequence, respectively. Here, ``height'' and ``width'' refer to the text tokens representing the spatial dimensions of the 2D image tokens. Following the special token ``<Image>'', the flattened image tokens are provided, with the length of ``height'' multiplied by ``width'' tokens. We utilize 1D RoPE~\cite{su2024roformer} consistent with the original language model instead of additional 2D positional encoding. 

\paragraph{Diffusion Decoder}
We chose to use a well-pretrained diffusion model as the visual decoder to reconstruct image pixels from discrete semantic tokens. Specifically, we added a linear layer to map the semantic embedding tokens to the feature channel dimensions of FLUX.1-dev~\cite{flux2024} and integrated it into the intermediate layer features. The diffusion decoder uses the semantic tokens extracted by the image tokenizer as input conditions and is trained with the objective of image reconstruction.

\subsection{Reinforcement Learning with GRPO}

To bridge the distribution gap between the semantic tokens used in the diffusion decoder during the training process and those generated by the autoregressive model, we employ reinforcement learning to provide comprehensive supervision for the autoregressive model. This guidance throughout the entire sampling process helps mitigate error propagation while ensuring that the output distribution of the autoregressive model aligns with the diffusion decoder's expectations. The overall pipeline for the reinforcement learning process is illustrated in Figure~\ref{fig:2}(a).

\subsubsection{GRPO Algorithm}

We adopt the Group Relative Policy Optimization (GRPO) \cite{shao2024deepseekmath} algorithm, which sidesteps the need for a separate critic network and thus reduces computational overhead. Specifically, for each text prompt $p\sim\mathcal{D}$, we generate a group of $G$ trajectories $\{o_1, o_2, ..., o_G\}$ using the previous policy $\pi_{\theta_{old}}$. These trajectories are then decoded by a fixed diffusion decoder to obtain the corresponding images $\{I_1, I_2, ..., I_G\}$, each of which is scored by our reward function to yield scalars $\{r_1,\dots,r_G\}$. We compute the advantages $A_i$ by normalizing this group of rewards, following the original GRPO procedure. The policy model $\pi_\theta$ is optimized through maximizing the following objective function:
\begin{equation}
\begin{split}
&\mathcal{J}_{GRPO}(\theta) = \mathbb{E}[{p \sim \mathcal{D}, \{o_i\}_{i=1}^{G} \sim \pi_{\theta_{old}}(\cdot|p)}] \\
&\ \ \ \frac{1}{G}\sum_{i=1}^{G} \left( \min\left( \frac{\pi_\theta(o_i|p)}{\pi_{\theta_{\mathrm{old}}}(o_i|p)}A_i, \mathrm{clip}\left(\frac{\pi_\theta(o_i|p)}{\pi_{\theta_{\mathrm{old}}}(o_i|p)}, 1-\epsilon, 1+\epsilon \right)A_i \right)-\beta\mathbb{D}_{KL}(\pi_\theta||\pi_{\theta_{ref}})\right),\\
\end{split}
\end{equation}
where $\epsilon$ and $\beta$ are hyper-parameters, $\pi_{\theta_{ref}}$ is a reference policy, and $\mathbb{D}_{\mathrm{KL}}$ is estimated using an unbiased estimator~\cite{Schulman20kl}. This formulation allows us to efficiently fine‐tune the policy by balancing reward maximization against divergence from a stable reference model.

\subsubsection{Reward System}

We construct a comprehensive reward system incorporating multiple specialized components, each designed to supervise distinct aspects of image generation quality. These diverse reward functions synergistically operate to provide comprehensive guidance during reinforcement learning, addressing critical dimensions such as aesthetic quality, text-image alignment, and text rendering accuracy. The individual reward signals are combined through a weighted aggregation mechanism to form the final reward score that guides the reinforcement learning optimization process. This multi-faceted approach ensures that our model learns to balance various quality criteria simultaneously, resulting in high-fidelity images that align with human expectations across multiple dimensions.

\paragraph{Human Preference Score}
We employ HPSv2~\cite{wu2023human} to assess aesthetic quality and human preference alignment. It effectively predicts human preferences for generated images and demonstrates robust generalization across various image distributions, making it an essential component for guiding our RL optimization toward aesthetically pleasing and human-aligned outputs.

\paragraph{Unified Reward Score}
While HPSv2 operates at 224$\times$224 resolution, our model specializes in high-resolution image generation. To address this limitation, we incorporate the Unified Reward~\cite{wang2025unified} model for human alignment assessment. 
This comprehensive reward aggregates multiple quality dimensions into a unified score, providing holistic feedback for reinforcement learning.

\paragraph{Text-Image Alignment Score}
To ensure semantic consistency between input prompts and generated images, we leverage the Qwen2.5-VL-32B~\cite{Qwen2.5-VL} vision-language model to compute an alignment reward. By harnessing the sophisticated image understanding capabilities of this VLM, we assess whether generated images accurately reflect the content described in the prompts. 
The alignment score quantifies the correspondence between textual descriptions and visual content, encouraging the generation of contextually relevant images while minimizing semantic hallucinations.

\paragraph{OCR Accuracy Score}
Text rendering accuracy represents a critical challenge in text-to-image generation. For prompts requiring textual generation within images, we implement an OCR-based reward that quantifies the fidelity of rendered text against ground truth. We adopt GOT-OCR2.0~\cite{wei2024general} and PaddleOCR~\cite{paddleocr2020} to jointly evaluate the images and calculate the accuracy score of text rendering.
This reward signal delivers critical guidance for enhancing text-to-image synthesis, enabling our model to reliably generate clear and precise textual content.

\section{Experiments}
\subsection{Training Data}
\paragraph{Pre-training}
The autoregressive model is pre-trained with a mixture of image generation and image understanding data. The image generation dataset contains open-source datasets including COYO-700M~\cite{kakaobrain2022coyo-700m}, DataComp-1B~\cite{datacomp}, and LAION-2B~\cite{schuhmann2022laion}. We design specific filters and collect approximately 200M high-quality images. Given the limited information and noise in original captions, we apply Qwen2.5-VL-72B model~\cite{Qwen2.5-VL} to create high-quality image-text pairs with dense captions. All images are resized to a short edge of 384 pixels and a maximum long edge of 1152 at native resolution ratio. The final image generation dataset contains 600B multimodal tokens after tokenization. Additionally, for image understanding task, we train with 59M data including data used in LLaVA-OneVision~\cite{li2024llava}, BLIP3-KALE~\cite{awadalla2024blip3}, Infinity-MM~\cite{gu2024infinitymmscalingmultimodalperformance}. We adopt the same resize strategy and produce about 100B multimodal tokens for image understanding tasks.

\paragraph{Supervised Finetuning}
After the large-scale pre-training, we perform a supervised finetuning stage. In this stage, we incorporate 30K high-quality data from BLIP3o-60k~\cite{chen2025blip3}, 30K synthetic text-to-image data, and high-quality pre-training tokens filtered from the pre-training dataset. We also mix image understanding data from LLaVA-NeXt~\cite{liu2024llavanext}, Cauldron~\cite{laurençon2024matters}, and Cambrian-1~\cite{tong2024cambrian1}. In summary, 1.5B tokens are trained in the SFT stage.

\paragraph{Reinforcement Learning}
\Ours leverages an on-policy GRPO algorithm, which requires only text prompts as inputs, while the images are generated by the model itself during training. To ensure that our training distribution comprehensively covers the capabilities we aim to enhance, we carefully curated a diverse reinforcement learning dataset from multiple sources. 
The dataset comprises 180K prompt samples from three distinct categories. First, we randomly sampled 80K cleaned prompts from the midjourney dataset~\cite{vivym24midjourney},  capturing the distribution of user requests to better align the training with real-world user preferences and expectations. Second, to enhance our model's text rendering capabilities during the reinforcement learning phase, we employed a bucket-based sampling strategy on our text-rich image data. Specifically, we sorted prompts by text length into different buckets and randomly extracted 10K samples from each bucket, ultimately collecting 50K text-rendering-focused prompts. Finally, to improve both aesthetic quality and instruction following abilities, we randomly sampled an additional 50K prompts from our natural image data.
This carefully balanced composition of 180K prompts spanning creative content generation, text rendering scenarios, and natural image descriptions provides a comprehensive training foundation for our reinforcement learning phase, enabling the model to excel across diverse user interaction scenarios. 

\begin{figure}[t]
  \centering
  \adjustbox{width=1.05\linewidth,center}{
        \includegraphics{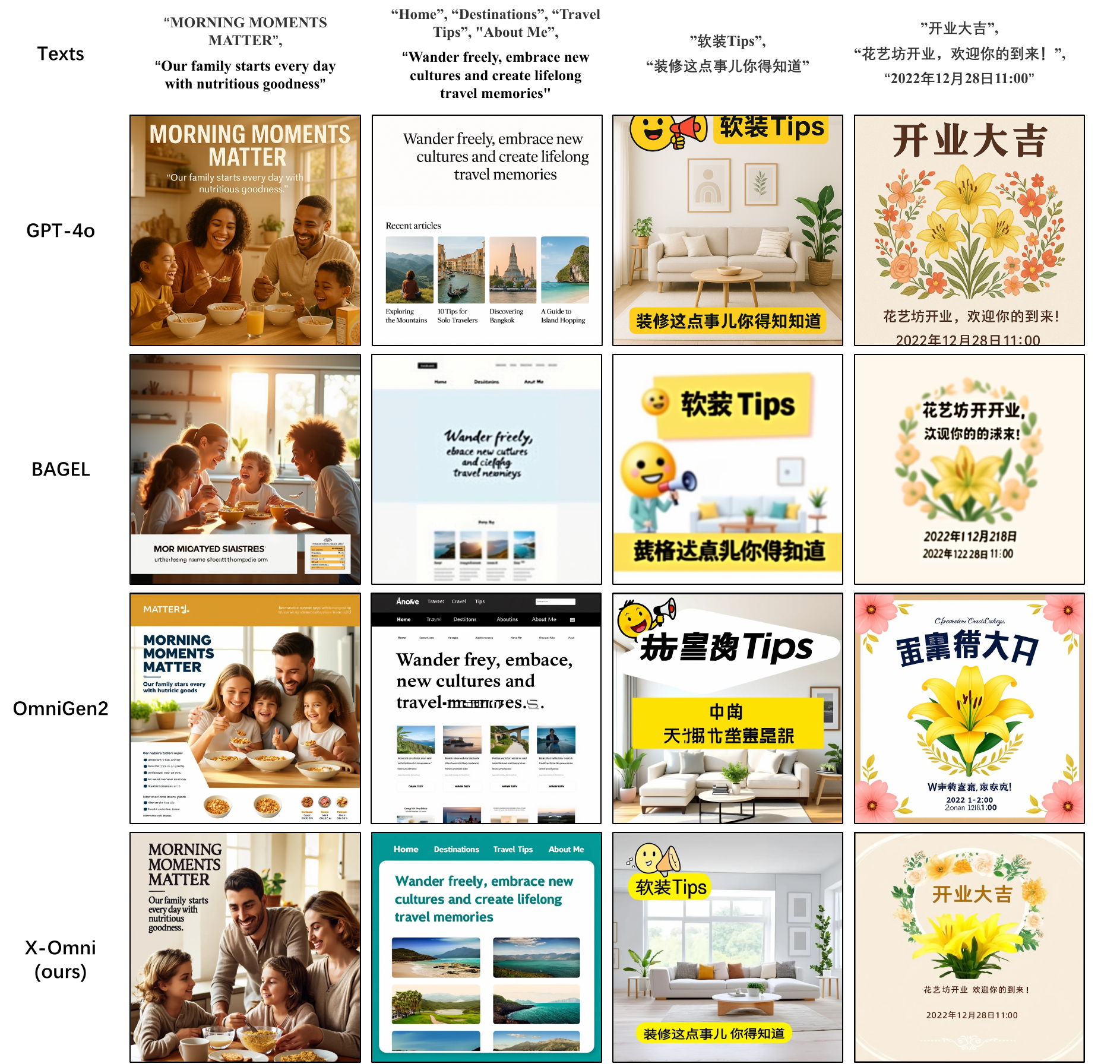}
  }
  \caption{Text rendering comparison with other unified multimodal models.}
  \label{fig:4}
\end{figure}

\subsection{Implementation details}
\paragraph{Pre-training}
We design a three-stage pre-training strategy for the autoregressive model. In the first stage, we only unfreeze the randomly initialized vision-specific blocks and vision token embeddings with generation data. In this stage, the training is conducted with a batch size of 256 and a sequence length of 16,384 for 10K steps, consuming 42B tokens in total. In the second stage, all components in the autoregressive model are set to be trainable, which are trained with a mixture of generation and understanding data. With the same batch size and sequence length as the first stage, the second stage is conducted with 150K steps and 629B tokens. Instead of a constant learning rate of 1$\times$10$^{-4}$ employed in the first two stages, we apply learning rate annealing in the third stage. The 42B high-quality tokens used in this stage are filtered from the pre-training dataset.

\paragraph{Supervised Finetuning} 
The supervised finetuning stage also trains only learnable parameters of the autoregressive model, using a batch size of 64 and a sequence length of 16384. In total, 1.5B tokens are trained in the SFT stage with the learning rate gradually reduced from $1 \times 10^{-5}$ to 0. 
During the SFT stage, the sequence packing remains consistent with pre-training, with the only differences being the data and the learning rate schedule.

\paragraph{Reinforcement Learning} In our reinforcement learning stage, we employ a learning rate of $1 \times 10^{-6}$ over 200 training steps. We utilize a batch size of 512, with each prompt generating 16 rollouts to ensure adequate exploration of the action space. Our loss function combines the standard policy gradient loss with a KL divergence term weighted at 0.01. This configuration provides a balance between learning stability and efficient policy improvement throughout the training process.
The Chinese text rendering model is derived by incorporating training on Chinese data at an intermediate checkpoint during the reinforcement learning stage.

\subsection{Results}

\begin{table}[t]
    \centering
    \renewcommand{\arraystretch}{1.25}
    \setlength{\tabcolsep}{0.45cm}
    \resizebox{0.85\textwidth}{!}{
    \begin{tabular}{lcccc}
        \toprule
        \multirow{2}{*}{\textbf{Method}} & \multicolumn{2}{c}{\textbf{OneIG-Bench}~\cite{chang2025oneig}} & \multicolumn{2}{c}{\textbf{LongText-Bench}} \\
        \cmidrule(r){2-5}
        & \textbf{English} & \textbf{Chinese} & \textbf{English} & \textbf{Chinese}  \\
        \midrule
        \textit{\textcolor{gray!60}{Gen. Only Models}} \\
        FLUX.1-dev~\cite{flux2024}
        & 0.523 & -     & 0.607 & 0.005 \\
        HiDream-I1-Full~\cite{cai2025hidream}
        & 0.707 & 0.205 & 0.543 & 0.024 \\
        Kolors 2.0~\cite{kolors2} 
        & 0.427 & 0.502 & 0.258 & 0.329 \\
        Seedream 3.0~\cite{gao2025seedream} 
        & 0.865 & 0.928 & 0.896 & 0.878 \\
        \midrule
        \textit{\textcolor{gray!60}{Unified Models}} \\
        Janus-Pro~\cite{chen2025janus} 
        & 0.001 & 0.148 & 0.019 & 0.006 \\
        BLIP3-o~\cite{chen2025blip3} 
        & 0.013 & 0.092 & 0.021 & 0.018 \\
        BAGEL~\cite{deng2025bagel} 
        & 0.244 & 0.365 & 0.373 & 0.310 \\
        OmniGen2~\cite{wu2025omnigen2} 
        & 0.680 & -     & 0.561 & 0.059 \\
        Show-o2~\cite{xie2025showo2}
        & 0.002 & -     & 0.006 & 0.002 \\
        GPT-4o~\cite{openai2025addendum} 
        & 0.857 & 0.650 & \textbf{0.956} & 0.619 \\
        \rowcolor{gray!15} \textbf{\Ours} 
        & \textbf{0.901} & \textbf{0.895} & 0.900 & \textbf{0.814} \\
        \bottomrule
    \end{tabular}
    }
    \vspace{10pt}
    \caption{Comparison of text rendering capability.}
    \label{tab:1}
    \vspace{-15pt}
\end{table}

After pre-training on image-text data, followed by supervised fine-tuning and reinforcement learning, we assess the performance of \Ours. The model demonstrates outstanding performance in complex instruction following and long text rendering. In this section, we evaluate on text rendering, text-to-image generation, and image understanding benchmarks. We also provide qualitative comparisons with other models to visually showcase \Ours's superior capabilities. Finally, we discuss two key findings about \Ours, including its ability to generate high-quality images without relying on classifier-free guidance.

\subsubsection{Text Rendering}
A notable advantage of \Ours lies in the ability to accurately render long texts, which is significantly enhanced through reinforcement learning. To quantitatively assess its performance on text rendering tasks, we select the Text Rendering task from OneIG-Bench~\cite{chang2025oneig} in both English and Chinese. This benchmark evaluates a composite score derived from three metrics: Edit Distance, Completion Rate, and Word Accuracy. The integration of these metrics provides a comprehensive evaluation of the model's proficiency in text rendering.

Furthermore, given the limited text lengths of the prompts from OneIG-Bench, we propose a novel benchmark, LongText-Bench, to fully evaluate our model's ability to accurately render long texts. This new benchmark consists of 160 meticulously designed prompts covering 8 different scenarios, which are specifically aimed at evaluating the capacity to precisely render long Chinese and English texts. Details about the setting of LongText-Bench can be found in Appendix~\ref{apx:bench}.

As shown in Table~\ref{tab:1}, for the English Text Rendering tasks from OneIG-Bench, \Ours significantly outperforms recent open-source unified works including BAGEL~\cite{deng2025bagel}, OmniGen2~\cite{wu2025omnigen2}, Show-o2~\cite{xie2025showo2}, and other proprietary models. For Chinese Text Rendering task from OneIG-Bench, \Ours surpasses most other models including GPT-4o~\cite{openai2025addendum} while achieving comparable performance with the specialized commercial text-to-image system Seedream 3.0~\cite{gao2025seedream}. In the English subpart of LongText-Bench, \Ours performs significantly better than other unified models, though it falls slightly behind GPT-4o. For Chinese long-text rendering evaluation, \Ours outperforms all other models by a large margin. Qualitative comparison examples are presented in Figure~\ref{fig:4} to further demonstrate that \Ours can precisely render long text according to instructions, while all other unified models (except GPT-4o) fail.

\begin{table}[t]
    \centering
    \renewcommand{\arraystretch}{1.3}
    \setlength{\tabcolsep}{0.32cm}
    \resizebox{0.9\textwidth}{!}{
    \begin{tabular}{@{\hspace{0.22cm}}lcccccc}
    \toprule
        \textbf{Method} & \textbf{Global} & \textbf{Entity} & \textbf{Attribute} & \textbf{Relation} & \textbf{Other} & \textbf{Overall } \\ 
        \midrule
        \textit{\textcolor{gray!60}{Gen. Only Models}} \\
        SDXL~\cite{podellsdxl} & 83.27 & 82.43 & 80.91 & 86.76 & 80.41 & 74.65  \\ 
        DALL-E~\cite{openai24dalle} & 90.97 & 89.61 & 88.39 & 90.58 & 89.83 & 83.50  \\ 
        SD3-medium~\cite{esser2024sd3} & 87.90 & 91.01 & 88.83 & 80.70 & 88.68 & 84.08  \\
        FLUX.1-dev~\cite{flux2024} & 82.10 & 89.50 & 88.70 & 91.10 & 89.40 & 84.00  \\ 
        \midrule
        \textit{\textcolor{gray!60}{Unified Models}} \\
        Emu3~\cite{wang2024emu3} & 85.21 & 86.68 & 86.84 & 90.22 & 83.15 & 80.60  \\ 
        Janus-Pro~\cite{chen2025janus} & 86.90 & 88.90 & 89.40 & 89.32 & 89.48 & 84.19  \\ 
        MetaQuery~\cite{pan2025transfer} & - & - & - & - & - & 82.05  \\ 
        BLIP3-o~\cite{chen2025blip3} & - & - & - & - & - & 81.60  \\ 
        UniWorld-V1~\cite{lin2025uniworld} & 83.64 & 88.39 & 88.44 & 89.27 & 87.22 & 81.38  \\ 
        Ovis-U1~\cite{wang2025ovisu1} & 82.37 & 90.08 & 88.68 & 93.35 & 85.20 & 83.72  \\ 
        Mogao~\cite{liao2025mogao} & 82.37 & 90.03 & 88.26 & 93.18 & 85.40 & 84.33  \\ 
        BAGEL~\cite{deng2025bagel} & 88.94 & 90.37 & 91.29 & 90.82 & 88.67 & 85.07  \\ 
        OmniGen2~\cite{wu2025omnigen2} & 88.81 & 88.83 & 90.18 & 89.37 & 90.27 & 83.57  \\ 
        Show-o2~\cite{xie2025showo2} & 89.00  & 91.78 & 89.96 & 91.81 & 91.64 & 86.14  \\ 
        GPT-4o$^\ast$~\cite{openai2025addendum} & 82.27 & 91.27 & 87.67 & 93.85 & 88.71 & 86.23 \\
        \rowcolor{gray!15} \textbf{\Ours} & 84.80 & 92.59 & 90.63 & 94.75 & 84.20 & \textbf{87.65} \\
        \bottomrule
    \end{tabular}
    }
    \vspace{10pt}
    \caption{Comparison of text-to-image generation performance on DPG-Bench~\cite{hu2024dpgbench}. $^\ast$: We generate only one image per prompt due to limited access to the official GPT-4o API. Additionally, several prompts from DPG-Bench are rejected by the official GPT-4o API, so the final results are calculated excluding these prompts.}
    \label{tab:2}
    \vspace{-10pt}
\end{table}

\begin{table}[t]
    \centering
    \renewcommand{\arraystretch}{1.3}
    \resizebox{0.9\textwidth}{!}{
    \begin{tabular}{lccccccc}
        \toprule
        \textbf{Method} & \textbf{Single} & \textbf{Two} & \textbf{Counting} & \textbf{Colors} & \textbf{Position} & \textbf{Color Attr.} & \textbf{Overall}  \\
        \midrule
        \textit{\textcolor{gray!60}{Gen. Only Models}} \\
        SDXL~\cite{podellsdxl} & 0.98 & 0.74 & 0.39 & 0.85 & 0.15 & 0.23 & 0.55  \\
        DALL-E~\cite{openai24dalle} & 0.96 & 0.87 & 0.47 & 0.83 & 0.43 & 0.45 & 0.67  \\
        SD3-medium~\cite{esser2024sd3} & 0.99 & 0.94 & 0.72 & 0.89 & 0.33 & 0.60 & 0.74  \\
        FLUX.1-dev~\cite{flux2024} & 0.98 & 0.93 & 0.75 & 0.93 & 0.68 & 0.65 & 0.82  \\
        \midrule
        \textit{\textcolor{gray!60}{Unified Models}} \\
        Emu3~\cite{wang2024emu3} & 0.99  & 0.81  & 0.42  & 0.80  & 0.49  & 0.45  & 0.66   \\
        Janus-Pro~\cite{chen2025janus} & 0.99  & 0.89  & 0.59  & 0.90 & 0.79  & 0.66  & 0.80   \\
        MetaQuery~\cite{pan2025transfer} & - & - & - & - & - & - & 0.80   \\
        BLIP3-o~\cite{chen2025blip3} & - & - & - & - & - & - & 0.84   \\
        UniWorld-V1~\cite{lin2025uniworld} & 0.99  & 0.93  & 0.81  & 0.89  & 0.74  & 0.71  & 0.84   \\
        Mogao~\cite{liao2025mogao} & 1.00  & 0.97  & 0.83  & 0.93  & 0.84  & 0.80  & \textbf{0.89}   \\
        BAGEL~\cite{deng2025bagel} & 0.98  & 0.95  & 0.84  & 0.95  & 0.78  & 0.77  & 0.88   \\
        OmniGen2~\cite{wu2025omnigen2} & 0.99 & 0.96 & 0.74 & 0.98 & 0.72 & 0.75 & 0.86  \\
        Show-o2~\cite{xie2025showo2} & 1.00  & 0.87  & 0.58  & 0.92  & 0.52  & 0.62  & 0.76   \\
        GPT-4o$^\ast$~\cite{openai2025addendum} & 0.99 & 0.92 & 0.85 & 0.92 & 0.75 & 0.61 & 0.84  \\ 
        \rowcolor{gray!15} \textbf{\Ours} & 0.98 & 0.95 & 0.75 & 0.91 & 0.71 & 0.68 & 0.83 \\
        \bottomrule
    \end{tabular}
    }
    \vspace{10pt}
    \caption{Comparison of text-to-image generation performance on GenEval~\cite{ghosh2023geneval}. $^\ast$: Results of GPT-4o are reported in \cite{yan2025gpt}.}
    \label{tab:3}
    \vspace{-5pt}
\end{table}

\subsubsection{Text-to-Image Generation}

\begin{figure}[h!]
  \centering
  \adjustbox{width=1.12\linewidth,center}{
        \includegraphics{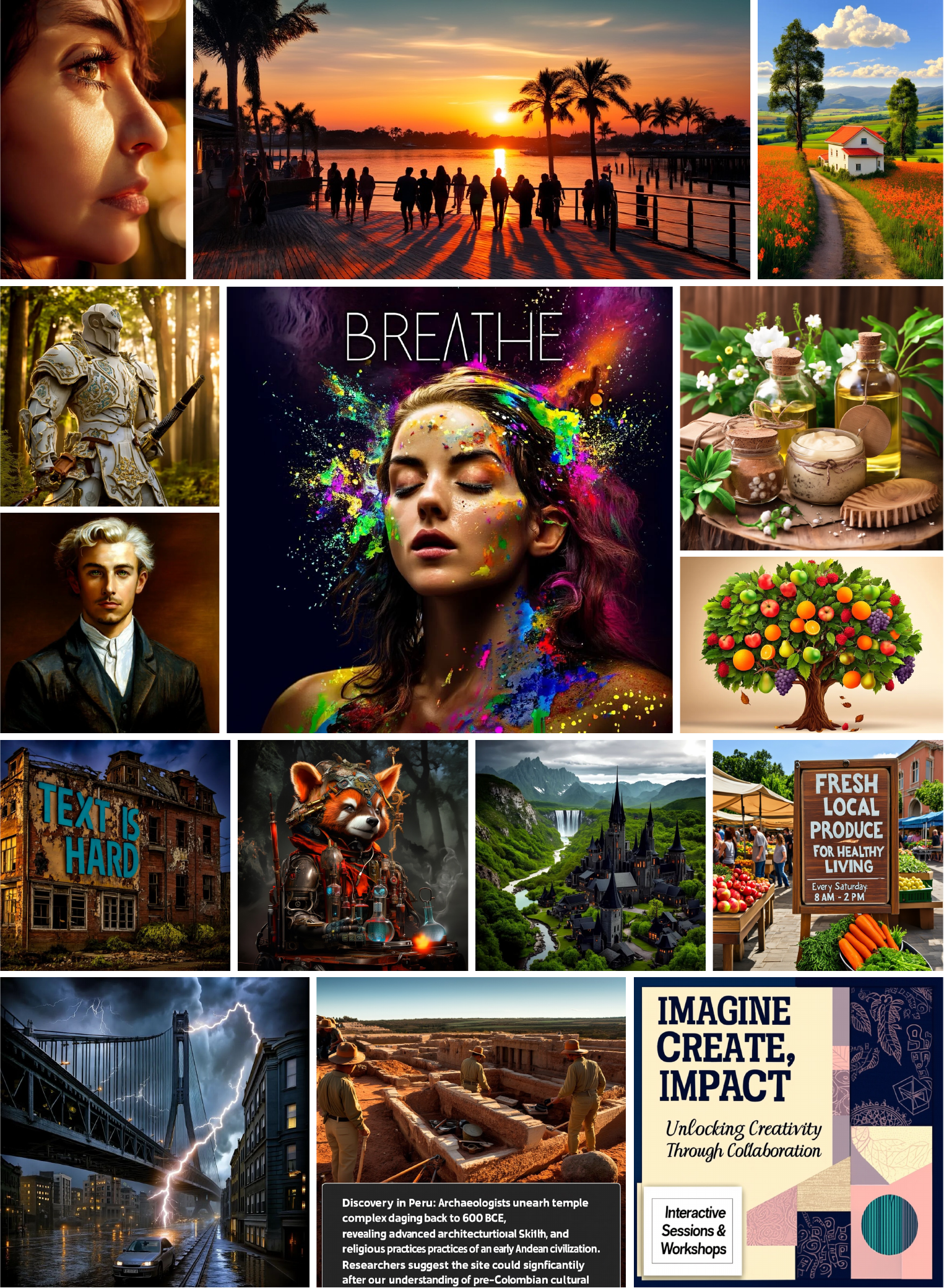}
  }
  \caption{Qualitative cases of \Ours.}
  \label{fig:5}
\end{figure}

We evaluate text-to-image generation on two widely recognized benchmarks: DPG-Bench~\cite{hu2024dpgbench} and GenEval~\cite{ghosh2023geneval}. Detailed results are shown in Table~\ref{tab:2} and Table~\ref{tab:3} respectively. Note that prompt rewriting is employed for GenEval evaluation. \Ours achieves state-of-the-art performance compared to recent unified models on DPG-Bench, and yields comparable results on GenEval. The excellent performance on these two benchmarks proves that \Ours can precisely follow complex generation instructions, which also demonstrates the universal benefits of reinforcement learning beyond text rendering. Furthermore, as illustrated in Figure~\ref{fig:5}, our method is capable of generating aesthetically pleasing images at arbitrary resolutions.

\begin{table}[t]
    \centering
    \renewcommand{\arraystretch}{1.3}
    \setlength{\tabcolsep}{0.28cm}
    \resizebox{1.0\textwidth}{!}{
    \begin{tabular}{lccccccc}
        \toprule
        \textbf{Method} & \textbf{\#LLM} & \textbf{POPE} & \textbf{GQA} & \textbf{MMB} & \textbf{SEED} & \textbf{DocVQA} & \textbf{OCRB}  \\
        \midrule
        \textit{\textcolor{gray!60}{Und. Only Models}} \\
        LLaVA-1.5~\cite{liu2024improved}
        & 7B   & 85.9 & 62.0 & 64.3 & 66.1 & -    & 318 \\
        LLaVA-NeXT~\cite{liu2024llavanext}
        & 7B   & 86.5 & 64.2 & 67.4 & 68.2 & 74.4 & 532 \\
        VILA~\cite{lin2023vila}
        & 7B   & 85.5 & 62.3 & 68.9 & 61.1 & -    & -   \\
        MiniCPM-Llama3-V 2.5~\cite{yao2024minicpmv}
        & 8B   & -    & -    & 77.2 & -    & 84.8 & 725 \\
        LLaVA-OneVision~\cite{li2024llava}
        & 7B   & -    & -    & 80.8 & 75.4 & 87.5 & 622 \\
        \midrule
        \textit{\textcolor{gray!60}{Unified Models}} \\
        Emu3~\cite{wang2024emu3}
        & 7B   & 85.2 & 60.3 & 58.5 & 68.2 & 76.3 & 687 \\
        Janus-Pro~\cite{chen2025janus}
        & 7B   & 87.4 & 62.0 & 79.2 & 72.1 & -    & 595   \\
        Mogao~\cite{liao2025mogao}
        & 7B   & 88.9 & 60.9 & 75.0 & 74.6 & -    & -   \\
        Show-o2~\cite{xie2025showo2}
        & 7B   & -    & 63.1 & 79.3 & 69.8 & -    & -   \\
        \rowcolor{gray!15} \textbf{\Ours}
        & 7B   & 89.3 & 62.8 & 74.8 & 74.1 & 88.6 & 704 \\
        \bottomrule
    \end{tabular}
    }
    \vspace{10pt}
    \caption{Results on image understanding benchmarks. The evaluation covers: POPE~\cite{li2023evaluating}; GQA~\cite{hudson2019gqa}; MMB: MMBench~\cite{liu2024mmbench}; SEED: SEEDBench-Img~\cite{li2024seed}; DocVQA~\cite{mathew2021docvqa}; OCRB: OCRBench~\cite{liu2024ocrbench}.}
    \label{tab:4}
    \vspace{-10pt}
\end{table}

\subsubsection{Image Understanding}

To evaluate image understanding performance, we present a comprehensive comparison across a wide range of public benchmarks in Table~\ref{tab:4}. Although our improvements primarily enhance generative capabilities, \Ours has achieved results comparable to the unified model Show-o2~\cite{xie2025showo2} on various image understanding benchmarks and has surpassed earlier works such as Emu3~\cite{wang2024emu3} and Janus-Pro~\cite{chen2025janus}. Notably, \Ours's results on OCRBench~\cite{liu2024ocrbench} have significantly exceeded those of the unified models Emu3 and Janus-Pro, as well as the specialized image understanding model LLaVA-OneVision~\cite{li2024llava}.

\subsection{Findings}

\begin{wrapfigure}{r}{0.4\linewidth}
  \centering
  \vspace{-25pt}
  \includegraphics[width=\linewidth]{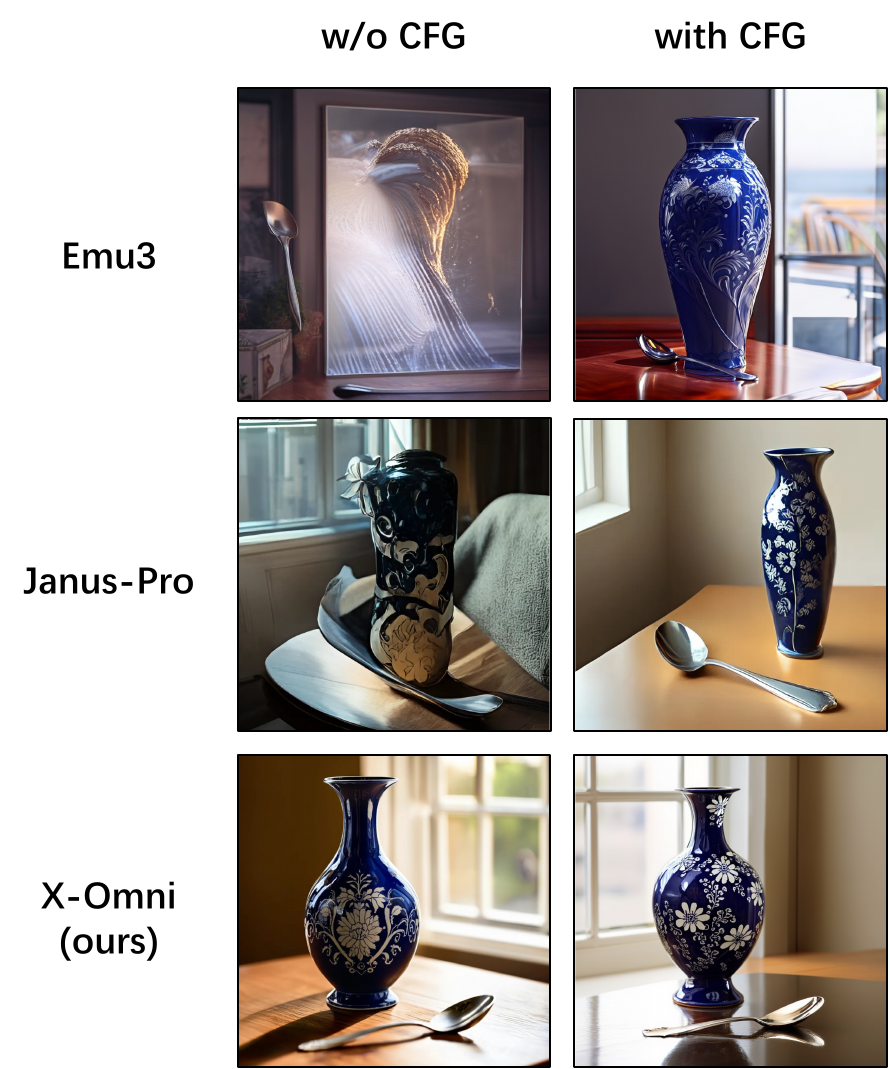}
  \caption{Comparison of dependency on classifier-free guidance (CFG).}
  \label{fig:6}
  \vspace{-15pt}
\end{wrapfigure}

\paragraph{\Ours Does Not Rely on Classifier-Free Guidance.}
A key observation is that our model can generate high-quality images without relying on classifier-free guidance (CFG) in the autoregressive component. Autoregressive image generation models, such as Emu3~\cite{wang2024emu3} and Janus-Pro~\cite{chen2025janus}, typically rely heavily on CFG to enhance the sampling of visual tokens. Figure~\ref{fig:6} compares the CFG dependencies and demonstrates that our method maintains consistently high generation quality, regardless of whether CFG is present. In contrast, the other autoregressive models experience a significant drop in generation quality when CFG is absent. \Ours operates independently of CFG, which not only lowers the computational cost of autoregressive inference but also indicates a higher level of consistency between the generation processes for visual and language tokens in our approach. Unless otherwise specified, all qualitative examples presented in this paper were generated without using CFG during autoregressive sampling.

\paragraph{RL outperforms SFT with Best-of-N sampling.}
As illustrated in Figure~\ref{fig:2}(b), RL training provides significant advantages to our image generation model, surpassing the best results achieved by the SFT model enhanced with best-of-N sampling. This stands in contrast to observations in language modeling, where the performance of SFT combined with best-of-N sampling often proves challenging to exceed through RL training. The disparity arises from two primary factors. First, SFT trains the AR and diffusion modules separately using ground-truth supervision, which results in performance degradation, whereas RL plays a crucial role in aligning these two modules. Second, unlike the sequential dependencies found in language, image features are inherently local and spatially complex. RL's holistic optimization harnesses rich, multifaceted information from various local regions within a single image, enabling highly efficient learning.

\section{Conclusion}

In this work, we propose using reinforcement learning to train an omni autoregressive model for unified image generation and understanding. We introduce \Ours, the first unified model capable of rendering long text, demonstrating the unique advantages of reinforcement learning. Additionally, our approach removes the dependency on classifier-free guidance during autoregressive sampling, highlighting the unified mechanisms of language and visual modeling within our framework.

\newpage

\bibliographystyle{unsrt}  
\bibliography{references}

\newpage
\appendix

\section{LongText-Bench Details}
\begin{figure}[t]
  \centering
  \adjustbox{width=1.0\linewidth,center}{
        \includegraphics{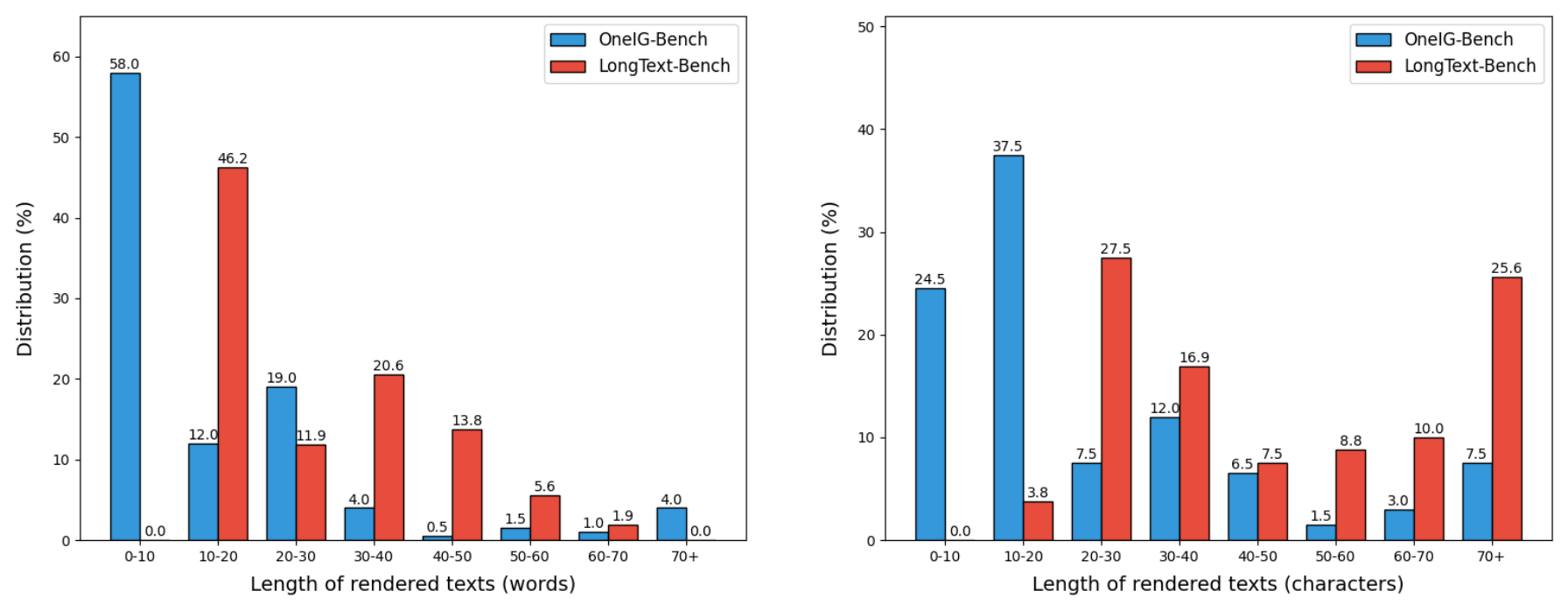}
  }
  \caption{Comparison between our proposed LongText-Bench and OneIG-Bench with respect to the length of rendered texts in English (left) and Chinese (right).}
  \label{fig:7}
  \vspace{-10pt}
\end{figure}

\paragraph{Benchmark Statistics}
Compared to current benchmarks, LongText-Bench focuses on evaluating the performance on rendering longer texts in both English and Chinese. To quantitatively illustrate the difference, we compare the distribution of the length of rendered texts in our benchmark with that in OneIG-Bench in Figure~\ref{fig:7}. For the English portion, the lengths of text contents from the ``short'' category of LongText-Bench are concentrated within the range of 10-30 words, while those in the ``long'' category predominantly fall within the range of 30-50 words. In the Chinese subset, the majority of prompts in the ``short'' category contain 20 to 40 characters, whereas the text rendered in prompts from the ``long'' category typically exceeds 60 characters in length. In comparison, the overall text length of our LongText-Bench exceeds that of OneIG-Bench, which better highlights \Ours's capability in long-text rendering tasks.

\label{apx:bench}

\paragraph{Prompt Construction}
Prompts in the LongText-Bench are meticulously curated through an automatic pipeline with manual post-review. In the first step, we define 8 common scenarios featuring text-rich contexts, including signboards, objects with labels, printed materials, web pages, slides, posters, captions, and dialogues. Subsequently, for each category, we instruct GPT-4o to generate 20 prompts for image generation comprising 10 prompts with short text contents and 10 prompts with longer text contents. After collecting the generated captions, we conduct manual review for each prompt and adjust the length of text contents to achieve a more balanced distribution. With this prompt construction pipeline, we finally curate a total of 160 prompts covering 8 categories for evaluating long text rendering tasks.

\paragraph{Evaluation Metric}
Following OneIG-Bench, we employ Qwen2.5-VL-7B~\cite{Qwen2.5-VL} as the OCR model to parse the generated texts. However, we observe that for long text scenarios, the OCR recognition results do not account for the relative order of different text segments. As a result, Edit Distance is not suitable for evaluating such cases. Therefore, we adopt Text Accuracy score as the final metric to assess the performance of text generation in LongText-Bench. Four images are generated for each prompt to mitigate the influence of random errors.

\end{document}